\documentclass{CVM}

\usepackage{graphicx}
\usepackage{comment}
\usepackage{amsmath,amssymb} 
\usepackage{color}
\usepackage{subfigure}
\usepackage{blkarray}
\usepackage{threeparttable}
\usepackage{multirow}
\usepackage{siunitx}
\usepackage{makecell}
\usepackage{colortbl}
\usepackage{fbox}
\usepackage{array}
\usepackage{booktabs}
\usepackage{wrapfig}
\usepackage{yhmath}
\usepackage{enumerate}
\usepackage{rotating}
\usepackage{diagbox}

\usepackage{url}

\usepackage[colorlinks,linkcolor=red,anchorcolor=blue,citecolor=blue]{hyperref}
\usepackage{bbm}

\usepackage{mathtools}

\usepackage{verbatim}
\usepackage{anyfontsize}

\renewcommand{\aboverulesep}{0pt}
\renewcommand{\belowrulesep}{0pt}

\usepackage{tikz}
\usepackage{xcolor} 
\newcommand{\yellowcircled}[2][yellow]{%
  \tikz[baseline=(c.base)]{
    \node[circle, fill=#1, draw=black, inner sep=0.1pt, minimum size=0.2em] (c) {\color{black}\strut #2};
  }%
}

\newcommand{\ie}{\textit{i.e.}}
\newcommand{\eg}{\textit{e.g.}}

\CVMsetup{
type      = {Research/Review Article},
doi       = {CVM.XXXX},
title     = {MAGE: View-guided Point Cloud Completion with Efficient Modality Alignment and Adaptive Geometry Enhancement},
author    = {Weize Quan$^{1,2}$, Zhengwei Wu$^{1,2}$, Kai Wang$^{3}$, and Dong-Ming Yan$^{1,2}$\cor{}\\
},
runauthor = {Weize Quan, et al.},
abstract  = {
  View-based point cloud completion aims to recover a complete 3D shape from a partial point cloud, guided by a single-view image. However, existing approaches often suffer from limited performance due to weak modality alignment and limited self-geometry enhancement. To overcome these challenges, we propose a unified geometry-aware framework that integrates efficient modality alignment and adaptive geometry enhancement, mainly to address cross-modal geometric inconsistency of view-guided point cloud completion. Specifically, we propose a geometry-aware modality alignment by integrating a shared self-attention Transformer and cross-modality reconstruction supervision, which aims to bring features of the image and point cloud close to each other in a shared latent space describing the 3D object. To enhance the perception of global shape and local geometric details, we propose an adaptive geometry-aware self-attention module, which simultaneously considers local geometry-aware attention computation and the spatially-variant feature fusion. In addition, we apply a geometry-perceptive anchor refinement module to reorganize the anchor points (representing a local region of the shape) under appropriate supervision, further boosting the completion performance of our method. Extensive experiments on both synthetic and real-world datasets demonstrate that our method achieves superior performance over existing approaches. Our code will be available at \url{https://github.com/weizequan/MAGE}.
},
keywords  = {Point cloud completion; Deep learning;  Modality alignment; Adaptive geometry-aware attention},
copyright = {The Author(s)},
}

\begin{document}

\maketitle

\enlargethispage{-3pt}
\begin{figure}[b] \vskip -4mm
\small\renewcommand\arraystretch{1.3}
\begin{tabular}{p{80.5mm}} \toprule\\ \end{tabular}
\vskip -4.5mm \noindent \setlength{\tabcolsep}{1pt}
\begin{tabular}{p{3.5mm}p{80mm}}
$1\quad $ & MAIS, Institute of Automation, Chinese Academy of Sciences, Beijing, China.\\
$2\quad $ & School of Artificial Intelligence, University of Chinese Academy of Sciences, Beijing, China. E-mail: qweizework@gmail.com; wuzhengwei2023@ia.ac.cn; yandongming@gmail.com.\\
$3\quad $ & GIPSA-lab, Univ. Grenoble Alpes, CNRS, Grenoble INP, 38000 Grenoble, France. E-mail: kai.wang@gipsa-lab.grenoble-inp.fr.\\
&\hspace{-5mm} Manuscript received: 2025-01-01; accepted: 2025-01-01\vspace{-2mm}
\end{tabular} \vspace {-3mm}
\end{figure}

\section{Introduction}
Recent advances in 3D sensing technology have significantly accelerated the development of 3D computer vision. Among various shape representations, point clouds have emerged as a widely adopted data format due to their compact storage and ability to preserve fine-grained geometric details. However, point clouds acquired from real-world sensors are often incomplete or degraded owing to inherent limitations such as self-occlusion, limited viewpoints, and reflective surfaces. These deficiencies hinder downstream applications, thus underscoring the need for robust methods to reconstruct complete and plausible point clouds from partial observations. Reliable completion is particularly critical for tasks such as 3D scene understanding~\citep{dai2018scan,luis20223D,shao2025iebins,tian2026bridging}, 3D reconstruction~\citep{dai2017shape,varley2017shape,sarmad2019rl,xu2023a}, industrial quality inspection~\cite{wang2026flow}, and autonomous
navigation~\citep{Pintore2024deep}.

\begin{figure*}[t]
  \centering
    \includegraphics[width=0.8\linewidth]{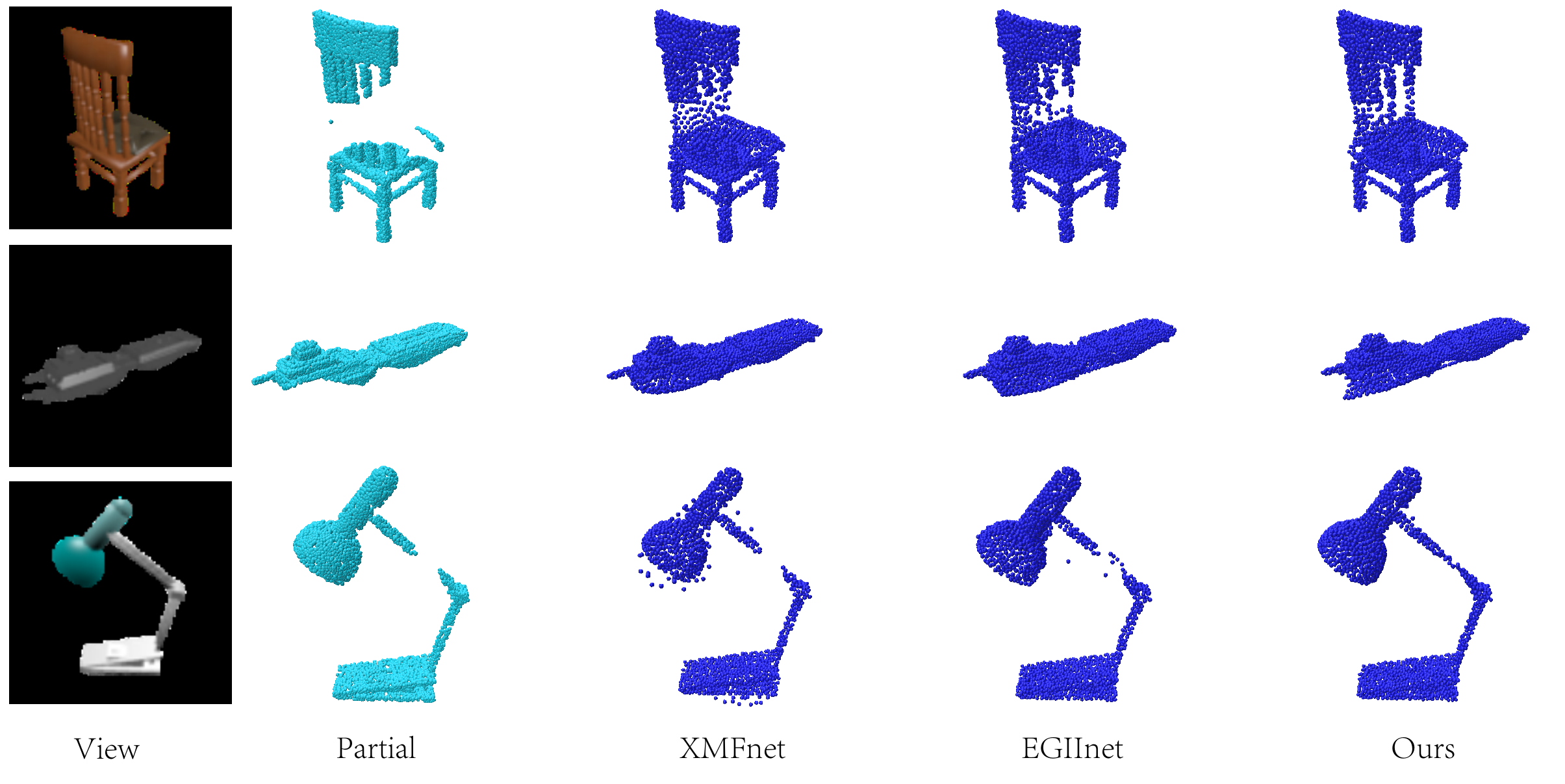}
    \caption{Several completed point clouds predicted by our method. Compared to recent methods (XMFnet and EGIInet), our method can recover the full global shape and more reasonable geometry details. }
    \label{fig:teaser}
\end{figure*}

Deep learning has substantially advanced point cloud completion, with progress spanning feature extraction, network architecture design, decoding strategies, and loss functions. For local point feature extraction, existing works have primarily explored kNN-based methods~\citep{pan2021var,Sipiran2022data}, EdgeConv-based approaches~\citep{zhu2023svd}, and attention mechanisms~\citep{su2023point,zhong2025point}. In terms of network architecture, both convolutional neural networks~\citep{wen2022pmp,zhang2023learning} and Transformers~\citep{pct2021,wu2024spac} have been widely employed. At the decoding stage, a variety of point generation strategies have been proposed, including folding-based methods~\citep{wen2020point}, coarse-to-fine pipelines~\citep{wang2022cascade,tang2022lake,Phong2023bilateral}, and hierarchical tree-structured decoders like TopNet~\citep{tchapmi2019topnet} and SPD~\cite{xiang2021snowflakenet,xiang2023snow}. Beyond the commonly used Chamfer distance, \cite{wu2021density} designed the density-aware Chamfer distance, \cite{lin2023hyper} proposed the hyperbolic Chamfer distance, and \cite{lin2023info} introduced contrastive Chamfer distance loss to further enhance completion quality.

The aforementioned point cloud completion methods primarily leverage 3D shape priors from a single-modality input—namely, the partial point cloud. However, due to the inherent sparsity and missing regions in such data, relying solely on this modality often leads to uncertainty and incomplete reconstructions. To address these limitations, recent research (\eg, ViPC~\citep{zhang2021view}, XMFnet~\citep{aiello2022cross}, CSDN~\citep{zhu2024csdn}, and EGIInet~\citep{xu2025explicit}) has focused on incorporating additional guidance from a single RGB image, which provides complementary visual and structural cues to enhance the completeness and accuracy of the reconstructed 3D shapes. Among existing view-guided point cloud completion approaches, however, there are two obvious challenges: (1) Insufficient modality alignment. Many methods encode two modalities of image and point cloud into latent feature vectors and rely on attention operations or feature transfer to align them. However, this process is largely implicit and lacks explicit geometric constraints. As a result, the model may fail to establish accurate cross-modal correspondences, leading to incomplete or structurally inconsistent reconstructions—for example, a missing tail wing of watercraft and a broken lamp stand in Fig.~\ref{fig:teaser}. A possible technical reason is that, without enforcing geometry-aware consistency (\eg, projection or correspondence constraints), the learned features are not sufficiently anchored to the underlying 3D structure, making it difficult to recover the complete shape or fine-grained consistency. (2) Limited self-geometry enhancement. While single-view images can provide complementary cues, existing methods still fall short in refining both global structure and fine-grained geometric details after image guidance. Prior approaches typically rely on either local graph-based aggregation in CSDN or spatially invariant global attention in XMFnet. However, these designs struggle to balance local detail preservation with global structural coherence. As shown in the chair back in Fig.~\ref{fig:teaser}, current view-guided point cloud completion methods (\eg, XMFNet and EGIInet) fail to accurately reconstruct thin structures and intricate local patterns. A possible reason is that local graph networks have limited receptive fields and are prone to noise accumulation, while conventional global attention tends to oversmooth geometric features due to the spatial location uniformity, leading to the loss of high-frequency details that are critical for precise shape recovery.

To address these challenges, we propose MAGE, an efficient Modality Alignment and adaptive Geometry Enhancement network for view-guided point cloud completion. Specifically, MAGE is composed of five main modules: a modality alignment module, an anchor refinement module, an image-guided cross-attention module, an adaptive geometry enhancement module, and a point cloud decoder. Firstly, we extract the token sequences of the image and input point cloud with respective tokenization processes and feed them into a shared Vision Transformer (ViT) to initially align with each other. Meanwhile, we explicitly match the image feature with the point cloud feature by reconstructing the point cloud from the input image using a decoder (with the same architecture as the point cloud decoder) and supervising this reconstruction with the ground-truth point cloud. Through the shared feature extractor and explicit geometric correspondence constraint, we enforce a better cross-modal alignment with good geometry-aware consistency. After tokenizing the input partial point cloud, we obtain a set of anchor points, each of which represents a local region of the input and encodes the location of the corresponding tokens. However, these anchors alone are insufficient to capture the complete shape. To address this limitation, we apply a geometry-perceptive anchor refinement module, which updates the anchor points guided by aligned point cloud features with supervised learning. Next, we inpaint the partial point cloud with the image guidance via a cross-attention module. To further improve the restoration of global shape and local details, we propose an adaptive geometry-aware self-attention module. This module combines the kNN-based attention and global self-attention to perceive the local and global simultaneously. In addition, we fuse the original point cloud feature and attentive features in a spatially-variant manner, considering the discrepancy between the valid and missing parts. This design jointly models local geometric relationships and global feature similarity, while adaptive feature fusion further mitigates the inherent oversmoothing effect of attention operations.
Finally, we apply a point cloud decoder to map the improved point cloud feature into the completed point cloud. With these novel designs, our MAGE can complete the partial point cloud with plausible global structure and reasonable local details, as shown in the last column of Fig.~\ref{fig:teaser}.

The main contributions of this work are as follows:
\begin{itemize}
    \item We propose an efficient modality alignment strategy that combines a shared self-attention Transformer with cross-modality reconstruction supervision.
    \item We introduce an adaptive geometry enhancement mechanism that integrates kNN-based local attention and global self-attention with spatially-variant feature fusion, refining both global structures and local geometry details.
    \item Extensive experiments on publicly available benchmarks demonstrate that our method consistently outperforms state-of-the-art approaches in both synthetic and real-world scenarios.
\end{itemize}

The remainder of this paper is structured as follows. Section~\ref{sec:related work} provides a brief review of existing point cloud completion methods. Section~\ref{sec:proposed method} details the network architecture and training objectives of the proposed approach. Section~\ref{sec:experiments} presents extensive experimental results and analysis. Finally, Section~\ref{sec:conclusions} concludes the paper and outlines potential directions for future research.

\begin{figure*}[ht]
\centering
\includegraphics[width=\linewidth]{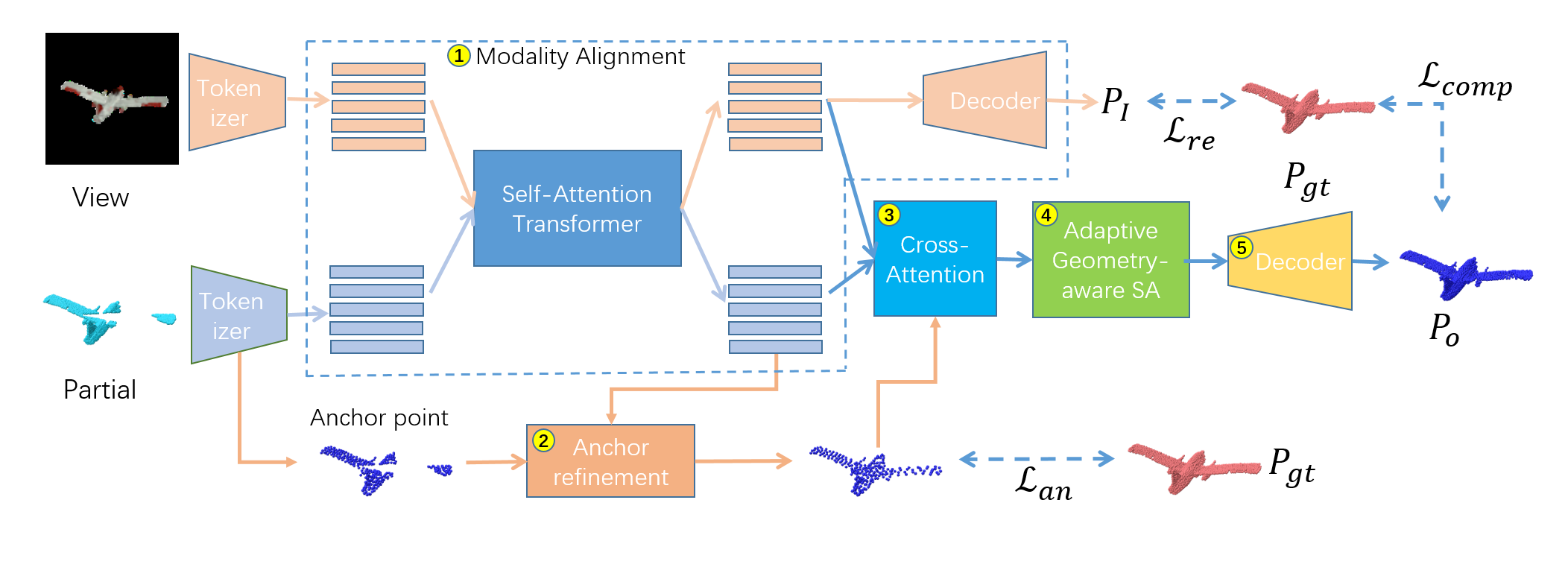}
\caption{The whole architecture of our proposed MAGE. It contains five components: a modality alignment module with a shared ViT and cross-modality reconstruction supervision (\yellowcircled{1}), an anchor refinement module with supervision (\yellowcircled{2}), an image-guidance cross-attention module (\yellowcircled{3}), an adaptive geometry-aware self-attention module (\yellowcircled{4}), and a point cloud decoder (\yellowcircled{5}). }
\label{fig:mage}
\end{figure*}

\section{Related Work}
\label{sec:related work}

\subsection{Point Cloud Completion}
An unordered point cloud is one of the most common representations of 3D shapes. Point cloud completion aims to recover the missing regions of an incomplete point cloud, producing a complete and plausible 3D geometry. A pioneering work, PCN~\citep{yuan2018pcn}, adopted an encoder-decoder framework, where the encoder was adapted from PointNet~\citep{qi2017pointnet} and the decoder combined a fully-connected generator~\citep{achlioptas2018learning} with FoldingNet~\citep{yang2018foldingnet}. Building upon this classical pipeline, subsequent works (\eg, MSN~\citep{liu2020morph}, GRNet \citep{xie2020grnet}, and PoinTr~\citep{yu2021pointr}) have introduced many improvements, including feature extraction mechanisms, novel network architectures, viewpoint-aware frameworks, refined decoding methodologies, etc. Some representative works are briefly reviewed below.

For the basic feature extraction process, several approaches (\eg, AtlasNet~\citep{Groueix2018a} and PF-Net~\citep{huang2020pfnet}) applied kNN-based local feature extraction inspired by PointNet and PointNet++~\citep{qi2017pointnet++}. Without using a max-pooling operation, \cite{wang2022softpool++} introduced soft-pooling to preserve more geometric information. Instead of depending on individual point-based convolution, \cite{yan2022fbnet,zhu2025pointsea} extracted the local point feature based on DGCNN~\citep{wang2019dynamic} with edge convolution (EdgeConv) to capture richer geometric relationships. To further enhance representational power, PointAttN~\cite{wang2024pointattn} introduced attention-based feature extraction techniques.

Transformers, with their strong sequence modeling capacity, have also been applied to point cloud completion. Works such as SDT~\cite{zhang2023point}, AnchorFormer~\citep{chen2023anchorformer}, AdaPoinTr~\cite{yu2023adapointr}, and CRA-PCN~\cite{rong2024cra} reformulated the point cloud completion task as a set-to-set translation problem. SeedFormer~\citep{zhou2022seedformer}, for example, utilized a Transformer encoder–decoder architecture and introduced patch seeds to incorporate local structural cues during decoding.

Point cloud acquisition is highly coupled with the scanning viewpoint; therefore, some researchers designed viewpoint-aware point cloud completion frameworks. Inspired by shadow volume, \cite{zhang2022shape} recovered the missing regions by optimizing the displacement of the partial scan given the camera location. ME-PCN~\cite{gong2021mepcn} explicitly encodes 3D emptiness via ray-based masking to jointly learn shape occupancy and non-occupancy, thereby preserving topology and boundaries. Fu et al.~\cite{fu2023vapcnet} proposed a viewpoint-aware point cloud completion method based on unsupervised viewpoint representation learning that leverages contrastive learning to implicitly capture unknown viewpoints, thereby enabling accurate and detailed point cloud completion. \cite{fu2024aednet} employed slot-attention-based global embeddings and a multiview-aware disentanglement across unit-sphere viewpoints to decompose point clouds into part-specific representations, enabling comprehensive geometric understanding for better completion performance.

At the decoding stage, diverse point generation strategies have been proposed. Wen et al. proposed a folding-based approach that reconstructs 3D point clouds by deforming a regular 2D grid. TopNet and SPD employed a hierarchical rooted tree structure to generate a dense point cloud progressively. \citep{yu2025facnet} adopted a sub-region recovery process with skip connections. Coarse-to-fine decoding methods~\citep{zhang2020detail} first predicted a coarse point cloud and then increased the density and details of the point clouds via expansion.

\subsection{View-guided Point Cloud Completion}

While traditional methods rely solely on incomplete point clouds and shape priors, view-guided point cloud completion (ViPC) incorporates additional cues from a single-view image to enhance reconstruction. ViPC first introduced this task, along with the ShapeNet-ViPC dataset for evaluation. 
Similar to ViPC, \cite{du2024cdpnet} first reconstructed the coarse point cloud from the image and then refined geometric details via multi-patch generators with AdaIN~\citep{adain2017}. \citep{luo2026rethinking} first reconstructed point clouds with a pre-trained image-to-3D model, and then corrected the features of partial observation with a hierarchical refinement process.
Unlike two-stage or multi-stage completion, we adopt a more natural one-stage framework with geometry-aware modality alignment and adaptive geometry completion. Instead of explicitly fusing the two modalities of image and point cloud in 3D space, XMFnet proposed latent-space fusion of image and point cloud features using stacked attention layers. CSDN introduced a coarse-to-fine framework consisting of shape transfer and dual refinement. The former transfers image information to the point cloud with a modified AdaIN, while the latter predicts the coordinate offsets by local refinement and global constraints. Based on the geometric nature of this completion task, EGIInet designed an explicit information interaction strategy with feature transfer loss, thereby unifying the encoding process of the two modalities to enhance modality alignment. 

Despite these efforts, existing attention-based ViPC approaches often adopt spatially uniform (or dense) attention mechanisms, which overlook the irregular distribution of point clouds (contrasting with the regular structure of images) and pay less attention to considering the discrepancy between the valid and missing regions of partial point clouds. Moreover, modality alignment remains a critical bottleneck: current strategies are mostly implicit and lack explicit geometric constraints. This highlights the need for more effective alignment mechanisms and adaptive feature integration strategies to fully exploit cross-modality cues for robust completion.

\subsection{Zero-Shot Point Cloud Completion}
Recently, there has been a growing trend toward zero-shot point cloud completion approaches that exploit powerful 2D and 3D generative priors. Kasten et al.~\cite{Kasten2023point} formulated point cloud completion as a conditional 3D generation problem guided by both a textual description and a partial point cloud. SDS-Complete relied on an SDF-based surface representation and performed test-time shape completion by coupling a pre-trained text-to-image diffusion model with the score distillation sampling (SDS) loss~\cite{poole2023dreamfusion}. Similarly, ComPC~\cite{huang2025compc} proposed a test-time framework that completes partial point clouds using a pre-trained 2D diffusion model. ComPC rendered partial point clouds via Gaussian Splatting and achieved shape completion by optimizing 3D Gaussians under the guidance of the 2D diffusion model. Li et al.~\cite{li2025genpc}proposed a depth-stepped point cloud completion framework that connects partial point clouds with image-to-3D generative models via depth prompting and preserves input geometry through adaptive pose- and scale-aware fusion.

\section{Our Method}
\label{sec:proposed method}

Fig.~\ref{fig:mage} illustrates the whole architecture of our network. It takes as input the partial point cloud $\mathbf{P}$ and a single-view image $\mathbf{I}$ and outputs a completed point cloud $\mathbf{P}_{o}$. We first apply the tokenization operations to extract the feature sequences of $\mathbf{P}$ and $\mathbf{I}$, respectively. Then, a shared ViT module is applied to align the two modalities initially. An image-to-point cloud reconstruction process is supervised and further enhances the alignment. Next, we design an anchor refinement block to regulate the anchor points to better represent the whole shape with appropriate supervision. Subsequently, we introduce a cross-attention module to complete the point cloud feature with the guidance of the image feature, and an adaptive geometry-aware self-attention module to enhance geometry details. Finally, a decoder is used to predict the final point cloud $\mathbf{P}_{o}$.

While individual components such as attention mechanisms or anchor-based optimization have been explored in prior work, our contribution lies in their principled integration within a unified geometry-aware framework. In particular, our method explicitly addresses the long-standing issue of cross-modal geometric inconsistency in view-guided point cloud completion, which has been largely overlooked in previous pipelines. To this end, we propose (i) explicit geometry-aware modality alignment, (ii) adaptive geometry-aware attention for feature enhancement, and (iii) geometry-perceptive anchor location refinement. These components are not ad hoc additions, but are tightly coupled to form a coherent framework that jointly improves global shape consistency and fine-grained geometric detail recovery, leading to superior completion performance.

\subsection{Tokenization}
\label{subsec:tokenization}

To achieve an initial alignment between the two modalities of images and point clouds using a shared ViT, we apply the prevailing tokenization processes. For the image modality, the input is divided into non-overlapping patches, which are then flattened into a token sequence $\mathcal{F}_I$, following the common practice in ViT-based models~\citep{vaswani2017attention}. For the point cloud modality, we draw inspiration from EGIInet and employ a farthest point sampling (FPS) strategy to extract representative features. Specifically, the point cloud is progressively downsampled using FPS, and at each stage, features are aggregated through a ball query–based neighborhood search. This process yields the feature set $\mathcal{F}_P$ corresponding to a fixed number of coarse points, referred to as anchors. Finally, positional embeddings are incorporated to encode the spatial information of these anchors.

\subsection{Efficient Modality Alignment}
In this work, we complete the partial point cloud with the guidance of an image. A natural approach is to map these two modalities to a shared latent space, as they describe the same object. The key challenge lies in effectively aligning point cloud and image features within this space. To this end, we propose an efficient modality alignment approach that integrates a shared self-attention Transformer with cross-modality reconstruction supervision. Specifically, we feed $\mathcal{F}_I$ and $\mathcal{F}_P$ into the shared Transformer to align the modalities of the point cloud and image initially and obtain aligned features $\mathcal{F}_I^{'}$ and $\mathcal{F}_P^{'}$. For modality-aware normalization, we respectively apply LayerNorm to the image token and the point cloud token after the shared ViT. To further enforce cross-modal consistency, the image feature $\mathcal{F}_I^{'}$ is decoded into a reconstructed point cloud $\mathbf{P}_{I}$ using an image decoder, which have the same architecture as the point cloud decoder described in Section~\ref{subsec:decoder}, which effectively narrows the gap between the two modalities in the latent space. The cross-modality reconstruction supervision is accomplished by introducing a reconstruction loss based on L2 Chamfer Distance (CD), and the corresponding formulation is written as:
\begin{equation}
    \mathcal{L}_{re} = \frac{1}{|\mathbf{P}_{I}|}\sum_{p \in \mathbf{P}_{I}}\underset{\bar{p} \in \mathbf{P}_{gt}}{\min}||p-\bar{p}||_2+\frac{1}{|\mathbf{P}_{gt}|}\sum_{\bar{p} \in \mathbf{P}_{gt}}\underset{p \in \mathbf{P}_{I}}{\min}||\bar{p}-p||_2,
    \label{equ:re}
\end{equation}
where $\mathrm{P}_{gt}$ is the ground truth, \ie, the complete point cloud.

\begin{figure}[t]
  \centering
    \includegraphics[width=\linewidth]{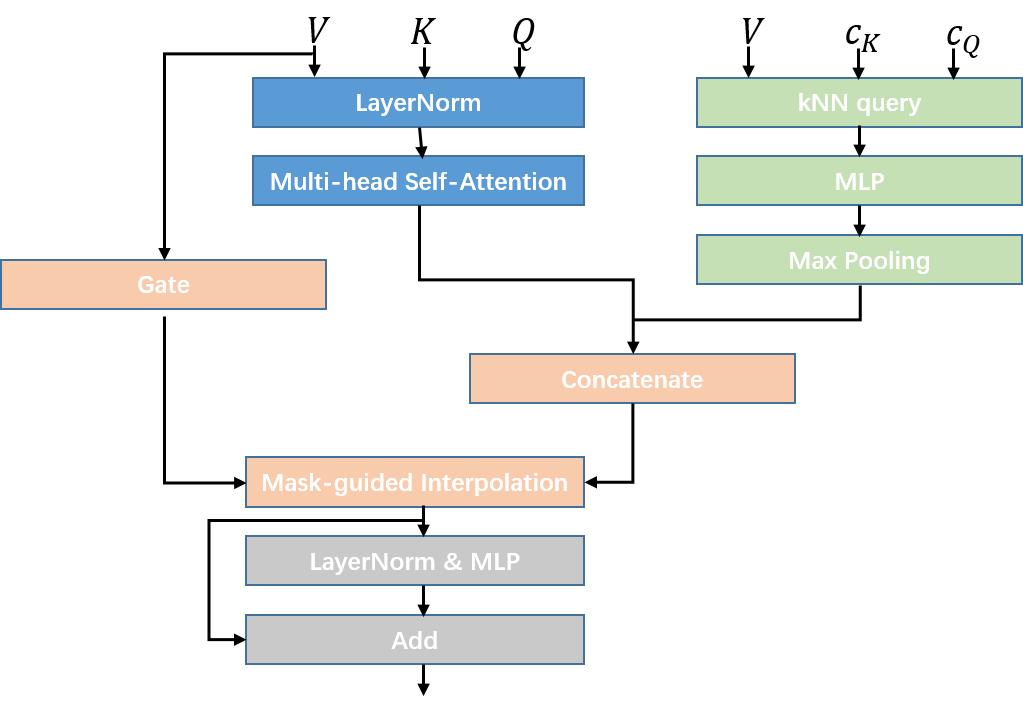}
    \caption{Overview of adaptive geometry-aware self-attention (AGSA). }
    \label{fig:AGSA}
\end{figure}

\subsection{Anchor Refinement}
In the tokenization of a partial point cloud, we apply an FPS-based downsampling to obtain anchor points. Due to the incompleteness of the input point cloud, the anchors only exist in the originally known regions and thus cannot represent the complete shape. In most previous methods, such as PoinTr, PointAttN, and AnchorFormer, only the global shape representation is utilized to predict the new positions of anchors. In contrast, we refine the spatial distribution of anchors by integrating both the global shape representation and the original anchor information. The key intuition is that the missing regions of a shape are partial and easier to infer, which leads to results that are more consistent with the original input. The ablation study in Section~\ref{subsec:ablation} also validates this design. Furthermore, this refinement process is supervised using the ground-truth shape. Specifically, we obtain the global shape representation by increasing the dimension of $\mathcal{F}_P^{'}$ and a max-pooling operation: 
\begin{equation}
    \mathcal{F}_g = \texttt{Maxpool}(\texttt{Conv1D}(\mathcal{F}_p^{'})),
\end{equation}
Then, we concatenate the global feature $\mathcal{F}_g$, point token features $\mathcal{F}_p^{'}$, and the coordinates of anchors $\mathbf{P}_{an}$, along the channel dimension. The refined anchor locations $\mathbf{P}_{an}^{re}$ are predicted with a cascade of linear layers with ReLU. This process is reported as:
\begin{equation}
    \mathbf{P}_{an}^{re} = \texttt{MLP}(\texttt{ReLU}(\texttt{MLP}(\texttt{Concat}(\mathcal{F}_g, \mathcal{F}_p^{'}, \mathbf{P}_{an})))).
\end{equation}

In addition, we introduce an anchor refinement loss building on the CD distance between $\mathbf{P}_{an}^{re}$ and $\mathbf{P}_{gt}$.
\begin{equation}
\begin{split}
    \mathcal{L}_{an} & =\frac{1}{|\mathbf{P}_{an}^{re}|}\sum_{p \in \mathbf{P}_{an}^{re}}\underset{\bar{p} \in \mathbf{P}_{gt}}{\min}||p-\bar{p}||_2 \\ & +\frac{1}{|\mathbf{P}_{gt}|}\sum_{\bar{p} \in \mathbf{P}_{gt}}\underset{p \in \mathbf{P}_{an}^{re}}{\min}||\bar{p}-p||_2,
    \label{equ:an}
\end{split}
\end{equation}

\subsection{Adaptive Geometry Enhancement}
\label{sec:AGSA}
After modality alignment and anchor refinement, we apply a cross-attention module to refine the point cloud feature with the guidance of the image feature. The cross-attention module is based on multi-head attention (MHA), and the formulation is written as:
\begin{equation}
   \mathcal{F}_P^{''} = \texttt{MHA}(\mathcal{F}_P^{'}, \mathcal{F}_I^{'}, \mathcal{F}_I^{'}).
\end{equation}
where we take $\mathcal{F}_P^{'}$ as query and $\mathcal{F}_I^{'}$ as key and value.

Although a single-view image can provide additional information, such as the object shape and part of the details, it cannot provide enough cues for the missing regions. In other words, after the cross-attention module, some regions are still not restored very well. Inspired by PoinTr and ARM3D~\citep{lan2022arm3d}, we introduce an adaptive geometry-aware self-attention module to further enhance shape restoration by jointly modeling local geometric correlations and global feature dependencies. In addition, we design a region-aware feature fusion mechanism that explicitly accounts for the discrepancy between missing and observed regions. This design not only improves feature discrimination across regions but also mitigates the potential oversmoothing effect of attention operations, leading to more accurate recovery of fine-grained structures. Technically, we combine the kNN-based attention with the traditional multi-head self-attention, as shown on the upper right of Fig.~\ref{fig:AGSA}. This is implemented as follows:
\begin{equation}
\begin{split}
    \mathcal{F}_s & = \texttt{MHA}(\texttt{LayerNorm}(V,K,Q)), \\
    \mathcal{F}_k & = \texttt{Maxpool}(\texttt{MLP}(\texttt{kNN}(V,c_K,c_Q))), \\
    \mathcal{F}_{sk} & = \texttt{MLP}(\texttt{Concat}(\mathcal{F}_s, \mathcal{F}_k)),
\end{split}
\end{equation}
where both $V$, $K$, $Q$ denote $\mathcal{F}_P^{''}$. For kNN-based attention, we search local feature sets of each point of coordinates $c_Q$ according to the key coordinates $c_K$. Then we fuse these feature sets via an MLP layer and max-pooling, like DGCNN.

To better merge the original point cloud feature and its hybrid attentive version, we design a new adaptive gate mechanism in a spatially variant manner. Specifically, we obtain a mask via an MLP and a sigmoid function. The final feature is obtained via mask-guided interpolation. The detailed process is denoted as:
\begin{equation}
\begin{split}
    M & = \texttt{Sigmoid}(\texttt{MLP}(V)), \\
    \hat{V} & = V * (1 - M) + \mathcal{F}_{sk} * M.
\end{split}
\end{equation}
The final feature is obtained via:
\begin{equation}
     \hat{V}^{'} = \hat{V} + \texttt{MLP}(\texttt{LayerNorm}(\hat{V})).
\end{equation}
The above process is shown in the bottom left of Fig.~\ref{fig:AGSA}.

\begin{table*}[ht]
\centering
\caption{Comparison of different methods on eight known categories ($\mathcal{L}_2-CD \times 10^3$). Lower is better.}
\begin{tabular}{lc *{8}{S[table-format=1.3]}}
\toprule
\multirow{2}{*}{Methods} & \multirow{2}{*}{Avg} & \multicolumn{8}{c}{Categories} \\
\cmidrule(lr){3-10}
 & & {Airplane} & {Cabinet} & {Car} & {Chair} & {Lamp} & {Sofa} & {Table} & {Watercraft} \\
\midrule
AtlasNet & 6.062 & 5.032 & 6.414 & 4.868 & 8.161 & 7.182 & 6.023 & 6.561 & 4.261 \\
FoldingNet & 6.271 & 5.242 & 6.958 & 5.307 & 8.823 & 6.504 & 6.368 & 7.080 & 3.882 \\
PCN & 5.619 & 4.246 & 6.409 & 4.840 & 7.441 & 6.331 & 5.668 & 6.508 & 3.510 \\
TopNet & 4.976 & 3.710 & 5.629 & 4.530 & 6.391 & 5.547 & 5.281 & 5.381 & 3.350 \\
PF-Net & 3.873 & 2.515 & 4.453 & 3.602 & 4.478 & 5.185 & 4.113 & 3.838 & 2.871 \\
MSN & 3.793 & 2.038 & 5.060 & 4.322 & 4.135 & 4.247 & 4.183 & 3.976 & 2.379 \\
GRNet & 3.171 & 1.916 & 4.468 & 3.915 & 3.402 & 3.034 & 3.872 & 3.071 & 2.160 \\
PoinTr & 2.851 & 1.686 & 4.001 & 3.203 & 3.111 & 2.928 & 3.507 & 2.845 & 1.737 \\
PointAttN & 2.853 & 1.613 & 3.969 & 3.257 & 3.157 & 3.058 & 3.406 & 2.787 & 1.872 \\
SDT & 4.246 & 3.166 & 4.807 & 3.607 & 5.056 & 6.101 & 4.525 & 3.995 & 2.856 \\
SeedFormer & 2.902 & 1.716 & 4.049 & 3.392 & 3.151 & 3.226 & 3.603 & 2.803 & 1.679 \\ 
\midrule
ViPC & 3.308 & 1.760 & 4.558 & 3.183 & 2.476 & 2.867 & 4.481 & 4.990 & 2.197 \\
CSDN & 2.570 & 1.251 & 3.670 & 2.977 & 2.835 & 2.554 & 3.240 & 2.575 & 1.742 \\
XMFnet & 1.443 & 0.572 & 1.980 & 1.754 & 1.403 & 1.810 & 1.702 & 1.386 & 0.945 \\
EGIInet & 1.211 & 0.534 & 1.921 & 1.655 & 1.204 & 0.776 & 1.552 & 1.227 & 0.802 \\
\textbf{Ours} & \textbf{1.094} & \textbf{0.526} & \textbf{1.615} & \textbf{1.580} & \textbf{1.083} & \textbf{0.692} & \textbf{1.391} & \textbf{1.122} & \textbf{0.743} \\
\bottomrule
\label{tab:cd}
\end{tabular}
\end{table*}

\begin{table*}[ht]
\centering
\caption{Comparison of different methods on eight known categories (Mean F-Score@0.001). Higher is better.}
\begin{tabular}{lc *{8}{S[table-format=1.3]}}
\toprule
\multirow{2}{*}{Methods} & \multirow{2}{*}{Avg} & \multicolumn{8}{c}{Categories} \\
\cmidrule(lr){3-10}
 & & {Airplane} & {Cabinet} & {Car} & {Chair} & {Lamp} & {Sofa} & {Table} & {Watercraft} \\
\midrule
AtlasNet & 0.410 & 0.509 & 0.304 & 0.379 & 0.326 & 0.426 & 0.318 & 0.469 & 0.551 \\
FoldingNet & 0.331 & 0.432 & 0.237 & 0.300 & 0.204 & 0.360 & 0.249 & 0.351 & 0.518 \\
PCN & 0.407 & 0.578 & 0.270 & 0.331 & 0.323 & 0.456 & 0.293 & 0.431 & 0.577 \\
TopNet & 0.467 & 0.593 & 0.358 & 0.405 & 0.388 & 0.491 & 0.361 & 0.528 & 0.615 \\
PF-Net & 0.551 & 0.718 & 0.399 & 0.453 & 0.489 & 0.559 & 0.409 & 0.614 & 0.656 \\
MSN & 0.578 & 0.798 & 0.378 & 0.380 & 0.562 & 0.652 & 0.410 & 0.615 & 0.708 \\
GRNet & 0.601 & 0.767 & 0.426 & 0.446 & 0.575 & 0.694 & 0.450 & 0.639 & 0.704 \\
PoinTr & 0.683 & 0.842 & 0.516 & 0.545 & 0.662 & 0.742 & 0.547 & 0.723 & 0.780 \\
PointAttN & 0.662 & 0.841 & 0.483 & 0.515 & 0.638 & 0.729 & 0.512 & 0.699 & 0.774 \\
SDT & 0.473 & 0.636 & 0.291 & 0.363 & 0.398 & 0.442 & 0.307 & 0.574 & 0.602 \\
SeedFormer & 0.688 & 0.835 & 0.551 & 0.544 & 0.668 & 0.777 & 0.555 & 0.716 & 0.786 \\
\midrule
ViPC & 0.591 & 0.803 & 0.451 & 0.512 & 0.529 & 0.706 & 0.434 & 0.594 & 0.730 \\
CSDN & 0.695 & 0.862 & 0.548 & 0.560 & 0.669 & 0.761 & 0.557 & 0.729 & 0.782 \\
XMFnet & 0.796 & 0.961 & 0.662 & 0.691 & 0.809 & 0.792 & 0.723 & 0.830 & 0.901 \\
EGIInet & 0.836 & 0.969 & 0.693 & 0.723 & 0.847 & 0.919 & 0.756 & 0.857 & 0.927 \\
\textbf{Ours} & \textbf{0.856} & \textbf{0.973} & \textbf{0.735} & \textbf{0.740} & \textbf{0.870} & \textbf{0.931} & \textbf{0.786} & \textbf{0.873} & \textbf{0.937} \\
\bottomrule
\label{tab:fscore}
\end{tabular}
\end{table*}

\begin{figure*}[ht]
\centering
\includegraphics[width=\linewidth]{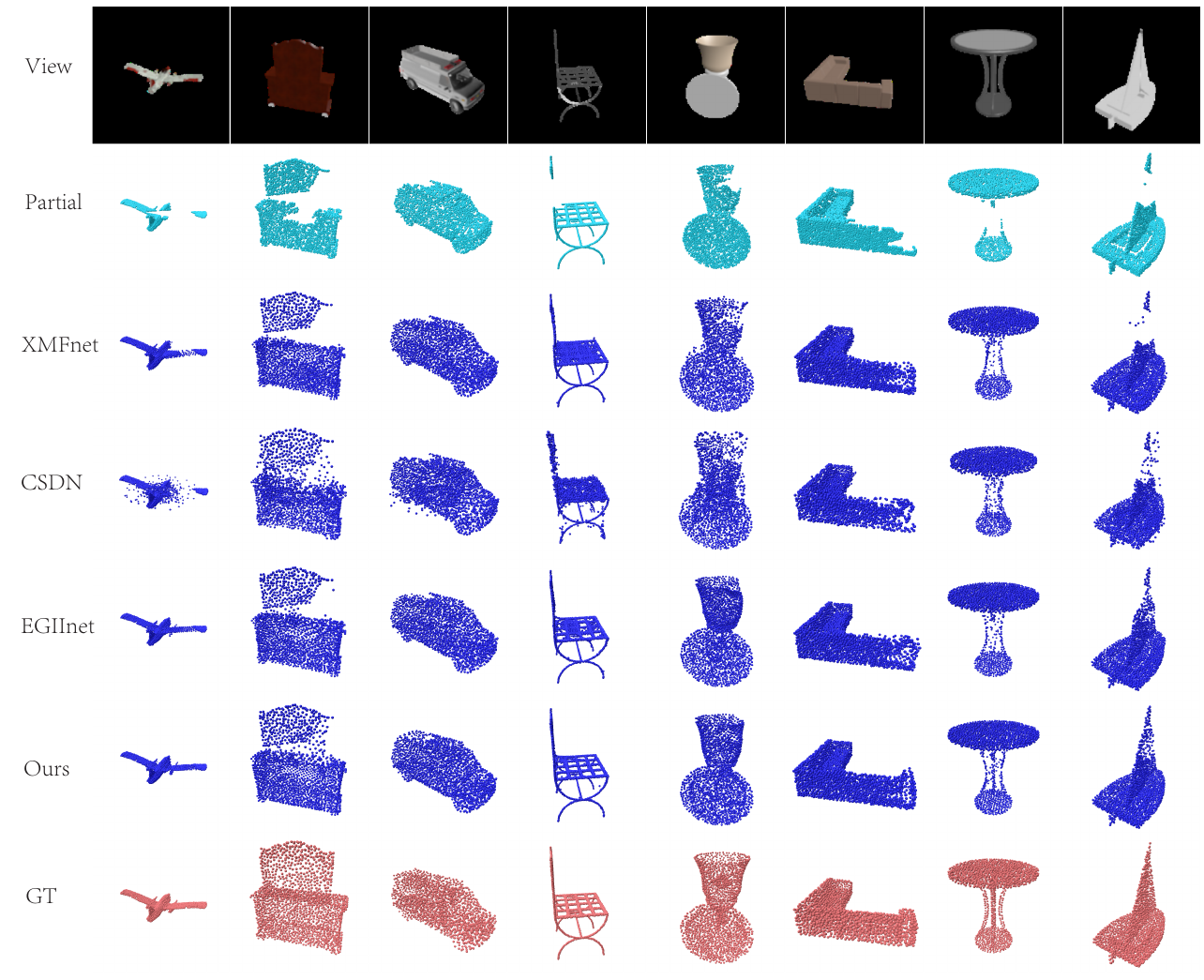}
\caption{Qualitative comparisons of our method with XMFnet, CSDN, and EGIInet on the ShapeNet-ViPC dataset with eight known categories. Each column corresponds to one category in Table~\ref{tab:cd}.}
\label{fig:main_result}
\end{figure*}

\begin{table*}[ht]
\centering
\caption{Comparison of different methods on unknown categories ($\mathcal{L}_2$-CD $\times 10^3$$\downarrow$ and Mean F-Score@0.001$\uparrow$).}
\begin{tabular}{l *{5}{S[table-format=1.3]@{\hspace{4pt}}S[table-format=1.3]}}
\toprule
\multirow{2}{*}{Methods} & \multicolumn{2}{c}{\textbf{Avg}} & \multicolumn{2}{c}{Bench} & \multicolumn{2}{c}{Monitor} & \multicolumn{2}{c}{Speaker} & \multicolumn{2}{c}{Cellphone} \\
\cmidrule(lr){2-3}\cmidrule(lr){4-5}\cmidrule(lr){6-7}\cmidrule(lr){8-9}\cmidrule(lr){10-11}
 & {CD} & {F-score} & {CD} & {F-score} & {CD} & {F-score} & {CD} & {F-score} & {CD} & {F-score} \\
\midrule
PF-Net & 5.011 & 0.468 & 3.683 & 0.584 & 5.304 & 0.433 & 7.663 & 0.319 & 3.392 & 0.534 \\
MSN & 4.684 & 0.533 & 2.613 & 0.706 & 4.818 & 0.527 & 8.259 & 0.291 & 3.047 & 0.607 \\
GRNet & 4.096 & 0.548 & 2.367 & 0.711 & 4.102 & 0.537 & 6.493 & 0.376 & 3.422 & 0.569 \\
PoinTr & 3.755 & 0.619 & 1.976 & 0.797 & 4.084 & 0.599 & 5.913 & 0.454 & 3.049 & 0.627 \\
PointAttN & 3.674 & 0.605 & 2.135 & 0.764 & 3.741 & 0.591 & 5.973 & 0.428 & 2.848 & 0.637 \\
SDT & 6.001 & 0.327 & 4.096 & 0.479 & 6.222 & 0.268 & 9.499 & 0.197 & 4.189 & 0.362 \\
\midrule
ViPC & 4.601 & 0.498 & 3.091 & 0.654 & 4.419 & 0.491 & 7.674 & 0.313 & 3.219 & 0.535 \\
CSDN & 3.656 & 0.631 & 1.834 & 0.798 & 4.115 & 0.598 & 5.690 & 0.485 & 2.985 & 0.644 \\
XMFnet & 2.671 & 0.710 & 1.278 & 0.862 & 2.806 & 0.677 & 4.823 & 0.556 & 1.779 & 0.748 \\
EGIInet & 2.354 & 0.750 & 1.047 &  0.902 &  2.513 & 0.716 & 4.282 & 0.591 &  1.575 &  0.792 \\
\textbf{Ours} & \textbf{2.306} & \textbf{0.764} & \textbf{1.009} & \textbf{0.910} & \textbf{2.441} & \textbf{0.738} & \textbf{4.272} & \textbf{0.599} & \textbf{1.503} & \textbf{0.808} \\
\bottomrule
\label{tab:unknow}
\end{tabular}
\end{table*}

\setlength{\tabcolsep}{4pt}
\begin{table*}[t]
\begin{center}
\caption{The completion performance of different variants of our network on eight known categories.}
\label{Tab:ab_main}
\footnotesize
\setlength{\tabcolsep}{4pt} \centering
  \begin{tabular}{c|cccccccc>{\centering\arraybackslash}p{1cm}|>{\centering\arraybackslash}p{1cm}}
    \hline
     Metric  & \multicolumn{8}{c}{$\mathcal{L}_2$-CD $\times 10^3$$\downarrow$} \\ \hline
     Categories  & Airplane & Cabinet & Car & Chair & Lamp & Sofa & Table & Watercraft \\ \hline 
     \textbf{Ours} & \textbf{0.526} & \textbf{1.615} & \textbf{1.580} & \textbf{1.083} & \textbf{0.692} & \textbf{1.391} & \textbf{1.122} & \textbf{0.743} \\ 
     w/o AR & 0.545 & 1.723 & 1.642 & 1.137 & 0.742 & 1.456 & 1.180 & 0.765  \\ 
     w/o AGSA  & 0.537 & 1.746 & 1.668 & 1.170 & 0.718 & 1.474 & 1.214 & 0.768  \\
     w/o CMR & 0.608 & 1.713 & 1.720 & 1.113  & 0.724 & 1.426 & 1.195 & 0.765   \\ \hline \hline
          Metric  & \multicolumn{8}{c}{Mean F-Score@0.001$\uparrow$}  \\ \hline
     Categories  & Airplane & Cabinet & Car & Chair & Lamp & Sofa & Table & Watercraft \\ \hline 
     \textbf{Ours}  & \textbf{0.973} & \textbf{0.735} & \textbf{0.740} & \textbf{0.870} & \textbf{0.931} & \textbf{0.786} & \textbf{0.873} & \textbf{0.937} \\ 
     w/o AR  & 0.970 & 0.719 & 0.722 & 0.858 & 0.923 & 0.774 & 0.864 & 0.933 \\ 
     w/o AGSA  & 0.971 & 0.714 & 0.720  & 0.852 & 0.928 & 0.769 & 0.857 & 0.932 \\
     w/o CMR & 0.959 & 0.721 & 0.711 & 0.864 & 0.924  & 0.780 & 0.865 & 0.933  \\ \hline
  \end{tabular}
\end{center}
\end{table*}

\begin{figure*}[ht]
\centering
\includegraphics[width=\linewidth]{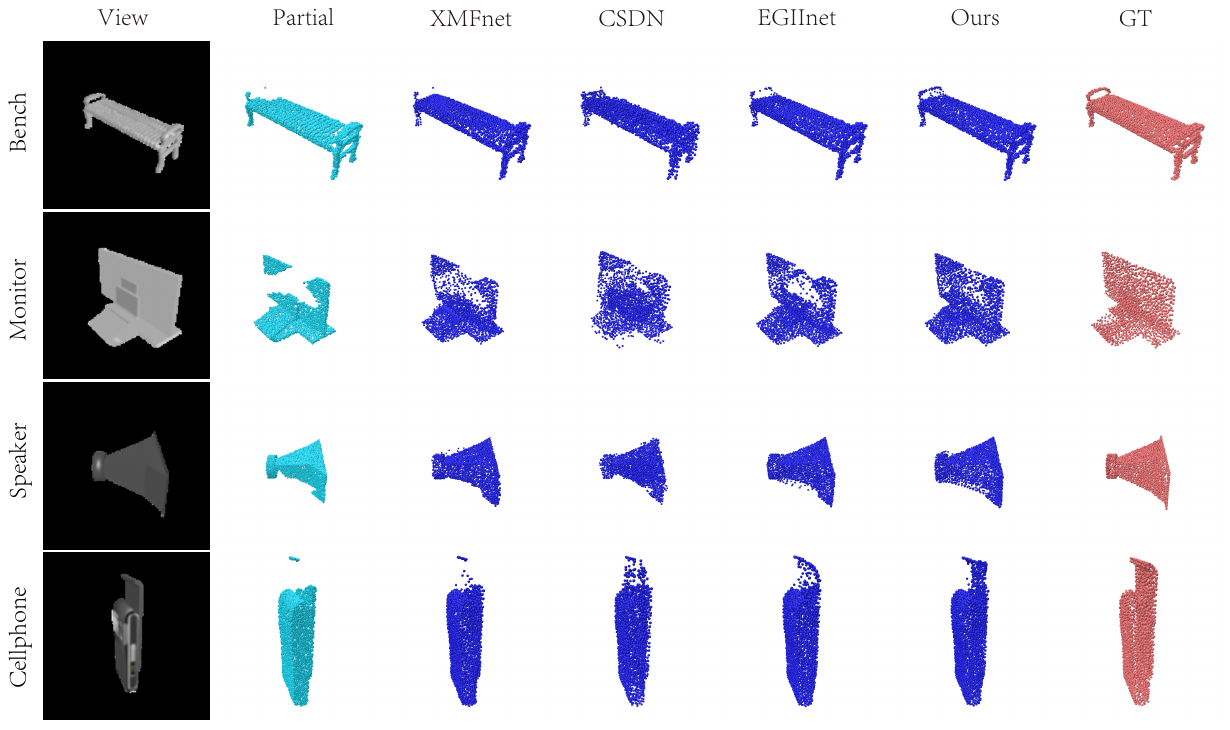}
\caption{Qualitative comparisons of our method with XMFnet, CSDN, and EGIInet on the ShapeNet-ViPC dataset with four unknown categories. Each row corresponds to one category in Table~\ref{tab:unknow}. }
\label{fig:comp_unseen}
\end{figure*}

\subsection{Decoder}
\label{subsec:decoder}

After obtaining the complete features, a decoder is required to map them into 3D point clouds with the desired resolution. For this purpose, we adopt a multi-branch upsampling network, following the design of XMFnet and EGIInet. Specifically, each branch is responsible for predicting a sub-region of the missing parts, and the predicted sub-regions are combined with downsampled versions of the original partial input to produce the final completed point cloud.

\subsection{Training Objectives}
To optimize the completed point cloud $\mathbf{P}_{o}$, we introduce the complete loss, which is defined as the Chamfer distance between $\mathbf{P}_{o}$ and its ground-truth value $\mathbf{P}_{gt}$:
\begin{equation}
    \mathcal{L}_{comp}=\frac{1}{|\mathbf{P}_{o}|}\sum_{p \in \mathbf{P}_{o}}\underset{\bar{p} \in \mathbf{P}_{gt}}{\min}||p-\bar{p}||_2+\frac{1}{|\mathbf{P}_{gt}|}\sum_{\bar{p} \in \mathbf{P}_{gt}}\underset{p \in \mathbf{P}_{o}}{\min}||\bar{p}-p||_2.
\end{equation}

The final training objective of our MAGE is a combination of the complete loss $\mathcal{L}_{comp}$, the cross-modality reconstruction loss $\mathcal{L}_{re}$ (Eq. (\ref{equ:re})), and the anchor refinement loss (Eq. (\ref{equ:an})). The formulation is written as:
\begin{equation}
\label{equ:all}
    \mathcal{L} = \mathcal{L}_{comp} + \alpha \cdot \mathcal{L}_{re} + \beta \cdot \mathcal{L}_{an},
\end{equation}
where $\alpha$ and $\beta$ are weighting parameters.

\section{Experiments}
\label{sec:experiments} 
In this section, we begin by detailing our experimental setup, including the datasets, implementation details, and evaluation metrics. We then assess the performance of our method through comparative experiments and generalization analyses, followed by additional results and comprehensive ablation studies.

\subsection{Experimental Settings}
\textbf{Dataset.} Following previous work, we evaluate our method on the ShapeNet-ViPC dataset, which contains 13 object categories and a total of 38,328 objects. Under 24 different viewpoints, each object accordingly produces 24 incomplete point clouds. In our work, we adopt the original train/test setting.

\textbf{Implementation Details.}
Our model is implemented using PyTorch 1.9.0 and trained on NVIDIA A6000 GPUs. We employ the Adam optimizer ($\beta_1 = 0.9$ and $\beta_2 = 0.999$) with a batch size of 64. The initial learning rate is set to 1e-3 and decayed by a factor of 10 every 32 epochs. Training is conducted for a total of 320 epochs. We set $\alpha = 0.1$ and $\beta = 0.5$ in all our experiments.

\textbf{Evaluation Metrics.}
Following the previous methods, we utilize two measures, \ie, CD (Chamfer Distance)~\citep{yang2019point} and F-Score~\citep{Maxim2019what}, as the quantitative metrics. The CD between the predicted point cloud $\mathcal{P}$ and the ground-truth point cloud $\mathcal{Q}$ is calculated by:
\begin{equation}
    d_{CD}(\mathcal{P},\mathcal{Q}) = \frac{1}{|\mathcal{P}|}\sum_{p \in \mathcal{P}}\underset{q \in \mathcal{Q}}{\min}||p-q||_2^2+\frac{1}{|\mathcal{Q}|}\sum_{q \in \mathcal{Q}}\underset{p \in \mathcal{P}}{\min}||q-p||_2^2.
    \label{equ:CD}
\end{equation}
The F-Score with a threshold ($\epsilon = 0.001$) is formulated as:
\begin{equation}
    F(\epsilon)=\frac{2P(\epsilon)R(\epsilon)}{P(\epsilon)+R(\epsilon)},
\end{equation}
where $P(\epsilon)$ and $R(\epsilon)$ denote respectively the precision and recall for the squared point cloud Euclidean distances less than the threshold $\epsilon$.

\subsection{Comparisons}
We compare our method with recently advanced view-guided point cloud completion approaches, including ViPC, XMFnet, CSDN, and EGIInet.

\textbf{Evaluation on known categories of ShapeNet-ViPC.} We first evaluate the completion performance of our method on known categories of ShapeNet-ViPC. Specifically, we train and test the model on one of the eight known categories (including airplane, cabinet, car, chair, lamp, sofa, table, and watercraft), and the corresponding numerical results are reported in Table~\ref{tab:cd} (in $\mathcal{L}_2-CD \times 10^3$) and Table~\ref{tab:fscore} (in Mean F-Score@0.001). From these two tables, we can see that our MAGE achieves the best performance on all metrics and categories. For instance, compared with the best performance of EGIInet, our approach reduces the average CD by 0.117 and improves the average F-Score by 0.020 (cf. the last two rows in the second column of ``Avg'' results in Table~\ref{tab:cd} and Table~\ref{tab:fscore}).

Fig.~\ref{fig:main_result} illustrates visual comparisons of different methods on eight categories (in columns). In some cases, XMFnet and CSDN cannot fill the original incomplete regions, \eg, the right wing of an airplane and the sail of a watercraft. Our method can recover these regions with reasonable structures. Compared to EGIInet, our method can predict the full structure (\eg, the dressing mirror of a cabinet in the second column and the armrest of a sofa in the sixth column) and better details (\eg, the chair seat in the fourth column and the table leg in the seventh column). These qualitative results demonstrate that our method can inpaint the partial point cloud with a complete structure and reasonable local details, owing to our efficient modality alignment and adaptive geometry enhancement.

\textbf{Results on unknown categories of ShapeNet-ViPC.} To further evaluate the generalization capability of our method, we test the trained model on 4 unseen categories, including bench, monitor, speaker, and cellphone. Specifically, we train our model on eight known categories, as shown in Table~\ref{tab:cd}, including cabinet, chair, sofa, and others. The corresponding numerical results are reported in Table~\ref{tab:unknow}. Our results consistently outperform all competitors in terms of CD and F-Score. Fig.~\ref{fig:comp_unseen} qualitatively compares our method with several advanced competitors on four unseen categories (in row). Among these methods, CSDN and XMFnet often fail to recover the complete structure, \eg, the monitor. EGIInet can provide relatively better results, while struggling to predict the geometry details, \eg, the phone cover. By contrast, our method can inpaint the whole shape (\eg, monitor and speaker) and the plausible local details (\eg, the bench armrest and the phone cover). These quantitative and qualitative comparisons demonstrate the better generalization capability of our method.

\begin{figure}[t]
\centering
\includegraphics[width=\linewidth]{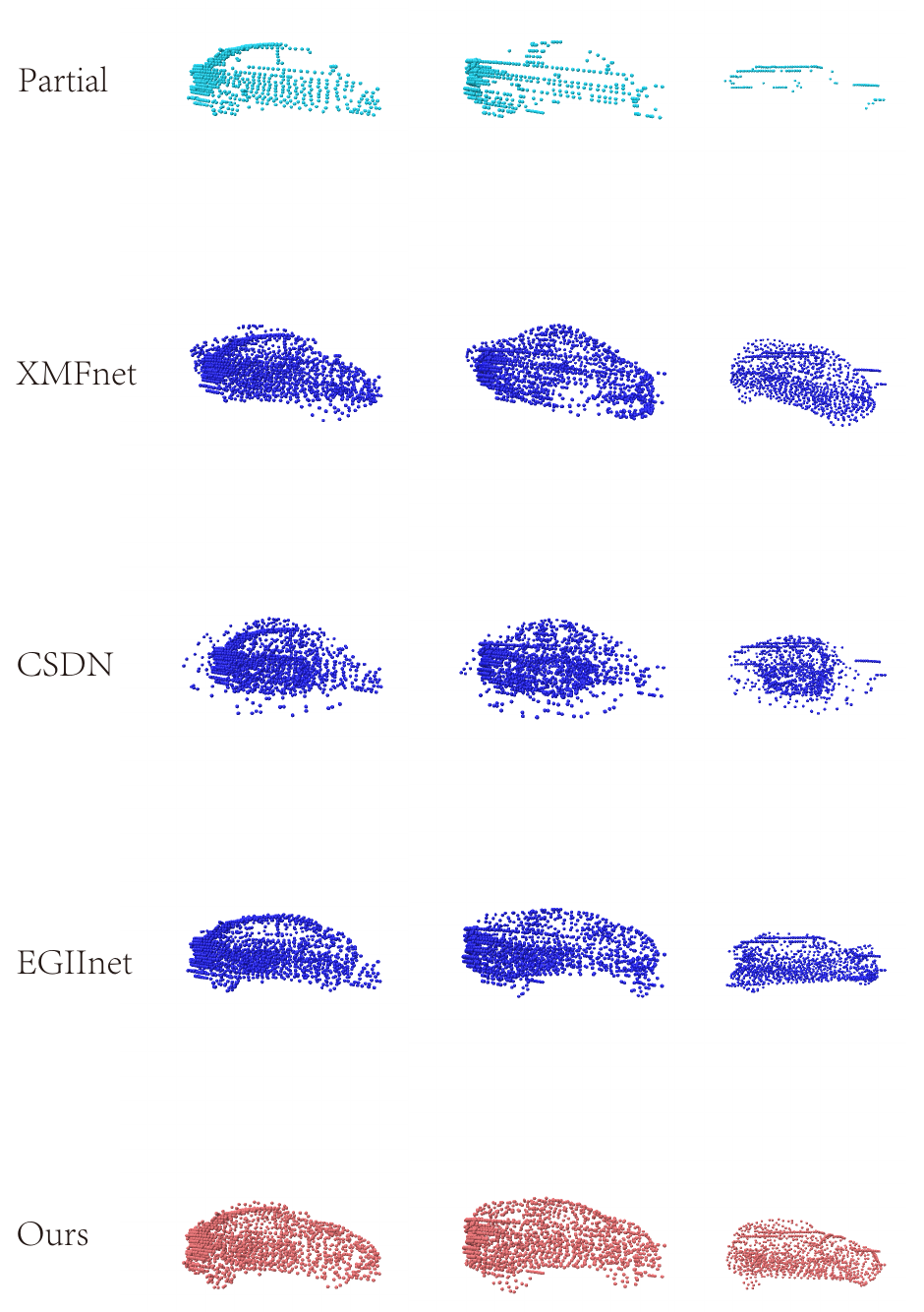}
\caption{Visual results of real-world scanned data.}
\label{fig:car_kitti}
\end{figure}

\textbf{Results on real-world scanned data.} In this part, we report the completed results on real-world scanned data. Specifically, we choose the KITTI cars shared by PCN, where incomplete car point clouds are segmented from the raw LiDAR scan in KITTI~\citep{kitti}. Following previous methods, our model is also trained on the category of cars in ShapeNet-ViPC and tested on KITTI cars. The quantitative results are shown in Fig.~\ref{fig:car_kitti}. Our method can better recover the local part (\ie, the car head in the first column) and the whole car structure (in the second column). When the original point cloud is heavily occluded, as shown in the third column, our method can predict the complete car. The visual results illustrate the completion performance of our method on real scenes.

\begin{figure*}[t]
\centering
\includegraphics[width=0.85\linewidth]{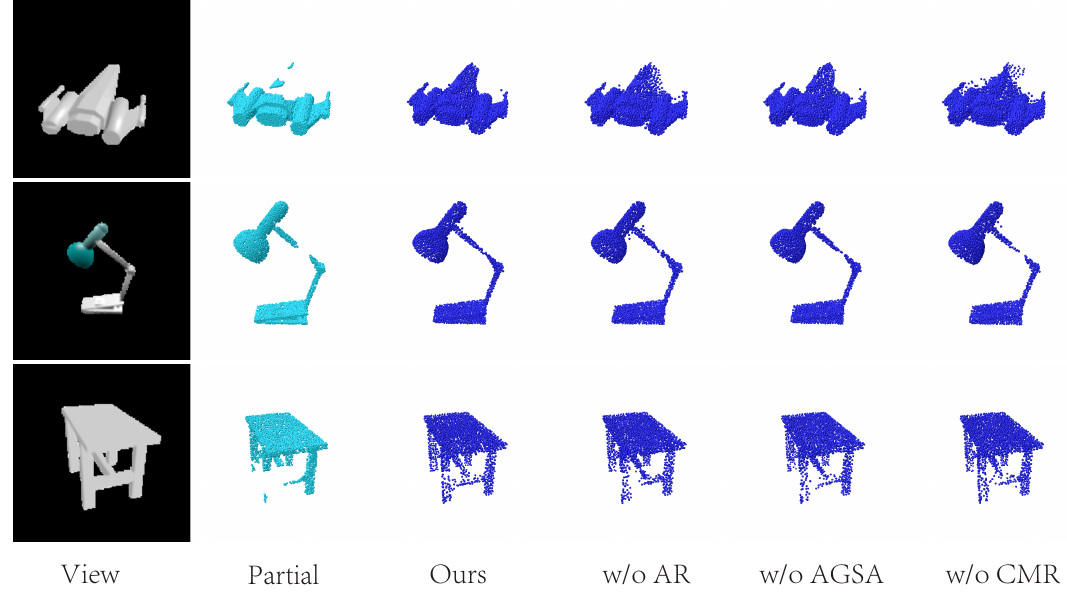}
\caption{Qualitative comparisons of our method with three variants on the categories of airplane, lamp, and table.}
\label{fig:comp}
\end{figure*}

\subsection{Ablation Studies}
\label{subsec:ablation}
\textbf{Network Design.} We evaluate our key components by comparing different variants of MAGE on eight known categories. The quantitative results are reported in Table~\ref{Tab:ab_main} and visual comparisons of several categories are shown in Fig.~\ref{fig:comp}. ``w/o AR'' refers to our network without anchor refinement;  ``w/o AGSA'' refers to our network without adaptive geometry-aware self-attention; ``w/o CMR'' refers to our network without cross-modality reconstruction supervision. In Table~\ref{Tab:ab_main}, we can see that removing each component of ``AR'', ``AGSA'', and ``CMR'' from our whole model, the completion performance of our method is remarkably degraded, especially for Cabinet, Car, and Table. When removing ``AR'', missing regions (\eg, the head of the plane in the fourth column of Fig.~\ref{fig:comp}) sometimes cannot be well inpainted with reasonable details. Without adaptive geometry-aware self-attention, local details are not completed well, \eg, the legs of the table in the fifth column. The inpainted performance is limited (\ie, the incomplete shape) due to the degraded modality alignment when removing the cross-modality reconstruction supervision. These quantitative and qualitative results consistently validate the effectiveness of our proposed components.

\setlength{\tabcolsep}{6pt}
\begin{table}[t]
\begin{center}
\caption{Ablation study on modality alignment.}
\label{Tab:modality_alignment}
\footnotesize
\setlength{\tabcolsep}{6pt} \centering
  \begin{tabular}{c|ccc>{\centering\arraybackslash}p{1cm}|>{\centering\arraybackslash}p{1cm}}
    \hline
     Metric  & \multicolumn{3}{c}{$\mathcal{L}_2$-CD $\times 10^3$$\downarrow$} \\ \hline
     Categories & Car & Chair & Sofa \\ \hline 
     Two separate ViTs & 1.961 & 1.123  & 1.447 \\ 
     Common alignment & 1.782 & 1.308  & 1.536 \\
     \textbf{One shared ViT (Ours)} & \textbf{1.720} & \textbf{1.113}  & \textbf{1.426} \\ \hline \hline
    Metric  & \multicolumn{3}{c}{Mean F-Score@0.001$\uparrow$}  \\ \hline
     Categories & Car & Chair  & Sofa \\ \hline 
     Two separate ViTs & 0.678 & 0.862  & 0.775 \\ 
     Common alignment & 0.697 & 0.827  & 0.757 \\
     \textbf{One shared ViT (Ours)} & \textbf{0.711} & \textbf{0.864}  & \textbf{0.780} \\ \hline
  \end{tabular}
\end{center}
\end{table}

\textbf{Modality Alignment.}
An important aspect of image-guided point cloud completion is to better align the image and point cloud modalities in the latent space. In our work, we combine a shared ViT and cross-modality reconstruction supervision. In this part, we experimentally analyze our design by comparing the completion performance of different settings: (1) two separate ViTs; (2) common modality alignment with attention. The corresponding results are reported in Table~\ref{Tab:modality_alignment}. Using two separate ViTs leads to performance degradation despite the increased number of parameters. A plausible reason is that independently learned encoders tend to produce misaligned feature spaces, making it difficult for downstream modules to effectively fuse information across modalities. Compared with the more conventional design, \ie, modality-specific encoders followed by an explicit alignment module (“Common alignment”), a shared ViT enforces a unified representation space, enabling feature alignment during encoding. This alignment is more stable and efficient, resulting in better cross-modal consistency and improved completion performance.

\textbf{Adaptive Geometry-aware SA.} To enhance the point cloud feature after guidance with the image modality, we propose an adaptive geometry-aware self-attention module in Section~\ref{sec:AGSA}. Compared to the common self-attention (denoted as ``Baseline''), the core idea of AGSA is to introduce the kNN-based attention (denoted as ``kNNA'') and mask-guided fusion (denoted as ``Fusion''). The numerical results are reported in Table~\ref{Tab:AGSA}. By modifying the structure of traditional self-attention with geometry-aware local attention and region-perceived feature fusion, the metrics of CD and F-Score are progressively improved. This indicates the superiority of our proposed adaptive geometry-aware self-attention.

\begin{figure}[t]
\centering
\includegraphics[width=0.85\linewidth]{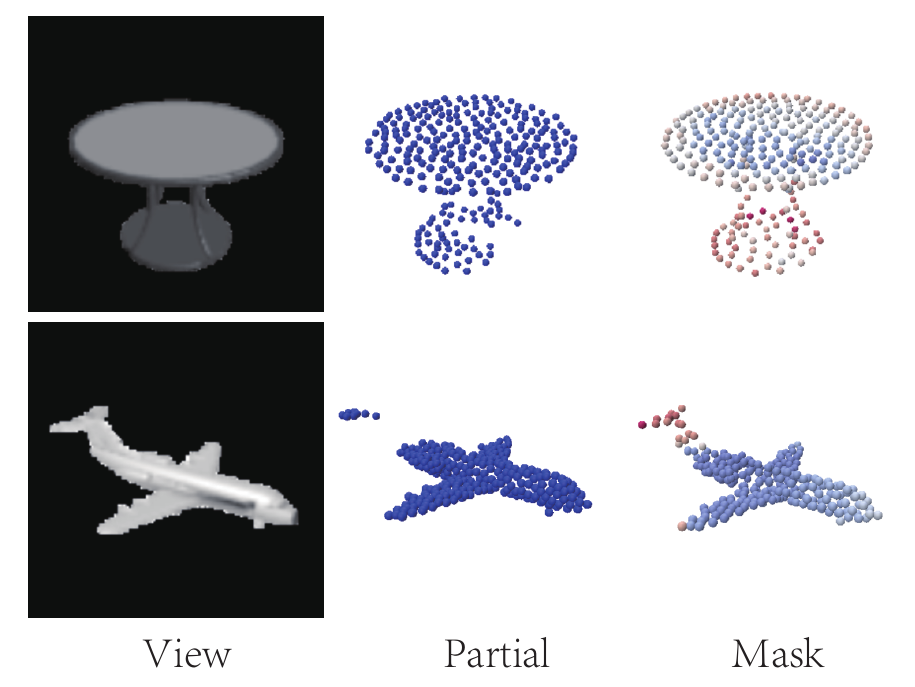}
\caption{Visualization of the mask in the AGSA module. ``Partial'' denotes the anchor points of the input partial point cloud, while ``Mask'' represents the rendered result of refined anchor points with fused weights. Larger weights are indicated in red, whereas smaller weights are shown in blue.  }
\label{fig:mask_vizalization}
\end{figure}

In our AGSA module, we introduce a mask-guided interpolation mechanism to enhance feature fusion. As described in Section~\ref{subsec:tokenization}, the input point cloud is tokenized using an FPS-based method, such that the mask weights correspond one-to-one with the refined anchor points. Fig.~\ref{fig:mask_vizalization} visualizes the mask in the AGSA module. It can be observed that the missing regions (by comparing the ``View'' and ``Partial'' columns) are assigned larger weights (red points in the ``Mask'' column), which are also different within the region. This indicates that our adaptive fusion mechanism can effectively perceive missing regions and better integrate features in a spatially variant manner.

\setlength{\tabcolsep}{4pt}
\begin{table}[t]
\begin{center}
\caption{The completion performance of different variants of AGSA.}
\label{Tab:AGSA}
\scriptsize
\setlength{\tabcolsep}{6pt} \centering
  \begin{tabular}{c||c|c||>{\centering\arraybackslash}p{1cm}|>{\centering\arraybackslash}p{1cm}}
    \hline
     Metric  & \multicolumn{2}{c||}{$\mathcal{L}_2$-CD $\times 10^3$$\downarrow$} & \multicolumn{2}{c}{Mean F-Score@0.001$\uparrow$}  \\ \hline
     Categories  &  Chair & Lamp & Chair & Lamp                \\ \hline \hline 
     Baseline    & 1.173 & 0.733 & 0.849 & 0.923 \\ 
     B+kNNA   & 1.156 & 0.716 & 0.853 & 0.927  \\
     \textbf{B+kNNA+Fusion}  & \textbf{1.083} & \textbf{0.692} & \textbf{0.870} & \textbf{0.931} \\ \hline
  \end{tabular}
\end{center}
\end{table}

\textbf{Anchor Refinement Module.}
In this work, we introduce a geometry-aware anchor refinement module to enhance the shape perception. To further evaluate the effect of our refined anchor mechanisms, we set several contrast configurations: (1) fixed anchor with a stronger decoder, \ie, TopNet, which is a carefully designed decoder architecture for point cloud completion; (2) fixed anchor with a simple decoder; (3) our original design (refined anchor with a simple decoder); (4) the new anchor point predicted only with the global shape information in PoinTr and a simple decoder. The corresponding numerical results are reported in Table~\ref{Tab:ab_anchor}. Compared with the combination of fixed anchors and a simple decoder (“Fixed+Simple”), introducing our anchor refinement strategy remarkably improves completion performance. When using fixed anchors, even with a more powerful decoder, the performance remains limited (“Fixed+Strong”). In addition, when only the global shape representation is used to predict new anchor locations—as in common approaches—the performance degrades (as shown by comparing “Refined+Simple” and “New+Simple”), which further validates our geometry-aware anchor location update. 

\setlength{\tabcolsep}{4pt}
\begin{table}[t]
\begin{center}
\caption{Ablation study on the anchor refinement module.}
\label{Tab:ab_anchor}
\footnotesize
\setlength{\tabcolsep}{4pt} \centering
  \begin{tabular}{c|cccc>{\centering\arraybackslash}p{1cm}|>{\centering\arraybackslash}p{1cm}}
    \hline
     Metric  & \multicolumn{4}{c}{$\mathcal{L}_2$-CD $\times 10^3$$\downarrow$} \\ \hline
     Categories   & Cabinet & Car & Sofa \\ \hline 
     Fixed+Strong  & 2.133 & 1.795 & 2.007 \\ 
     Fixed+Simple   & 1.723 & 1.642 & 1.456\\
     \textbf{Refined+Simple (Ours)} & \textbf{1.615} & \textbf{1.580} & \textbf{1.391} \\ 
     New+Simple & 1.941 & 2.661 & 1.603 \\ \hline 
      \hline
          Metric  & \multicolumn{4}{c}{Mean F-Score@0.001$\uparrow$}  \\ \hline
     Categories  & Cabinet & Car & Sofa \\ \hline 
     Fixed+Strong  & 0.692 & 0.724  &  0.693\\ 
     Fixed+Simple  & 0.719 & 0.722  & 0.774 \\
     \textbf{Refined+Simple (Ours)} & \textbf{0.735} & \textbf{0.740} & \textbf{0.786} \\ 
     New+Simple & 0.688 & 0.606 & 0.751 \\ \hline
  \end{tabular}
\end{center}
\end{table}

 \textbf{Reconstruction Loss.} In this work, we strengthen modality alignment by introducing a cross-modal prediction module supervised with a reconstruction loss (Eq. (\ref{equ:re})). To deeply analyze the contribution of this reconstruction loss, we conduct an ablation study on the hyperparameter $\alpha$ in Eq. (\ref{equ:all}). Specifically, we evaluate the inpainting performance under different values of $\alpha$, and the corresponding results are presented in Table \ref{Tab:re}. Compared with the rows of $\alpha = 0.0$ and $\alpha = 0.1$, incorporating the reconstruction loss improves completion performance. However, as $\alpha$ increases further, the performance degrades. A possible explanation is that a larger $\alpha$ enforces the image features to overly resemble the point cloud representation, yielding more similar reconstructed results to the point cloud. This excessive feature transfer may impair the integrity of the original image features, which are essential for recovering missing regions.

\setlength{\tabcolsep}{4pt}
\begin{table}[t]
\begin{center}
\caption{The ablation experiment of the reconstruction loss.}
\label{Tab:re}
\footnotesize
\setlength{\tabcolsep}{6pt} \centering
  \begin{tabular}{c||c|c||>{\centering\arraybackslash}p{1cm}|>{\centering\arraybackslash}p{1cm}}
    \hline
     $\alpha$  & \multicolumn{2}{c||}{$\mathcal{L}_2$-CD $\times 10^3$$\downarrow$} & \multicolumn{2}{c}{Mean F-Score@0.001$\uparrow$}  \\ \hline
     Categories  & Chair & Lamp & Chair & Lamp                \\ \hline \hline
     0.0 & 1.113 & 0.724 & 0.864 & 0.924 \\
     \textbf{0.1} & \textbf{1.083} & \textbf{0.692} & \textbf{0.870} & \textbf{0.931} \\ 
    0.2  & 1.110 & 0.744   & 0.864 & 0.922  \\
     0.3  & 1.128  & 0.718  & 0.860 & 0.925    \\ \hline
  \end{tabular}
\end{center}
\end{table}

\subsection{Failed Cases and Limitations}
\begin{figure}[t]
  \centering
    \includegraphics[width=0.85\linewidth]{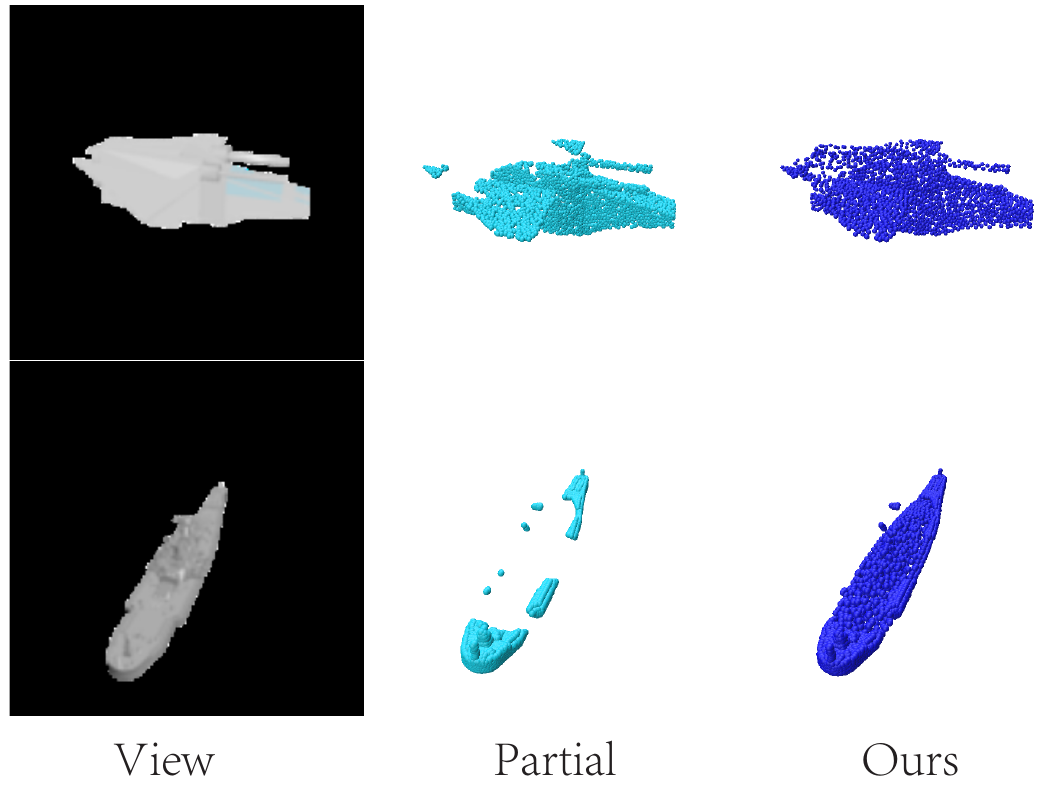}
    \caption{Two failed examples of our method. }
    \label{fig:failed_case}
\end{figure}

Our method may fail to restore fine-grained details when the provided reference images are blurry due to viewpoint limitations or low resolution, as illustrated in Fig.~\ref{fig:failed_case}. Although the proposed image-guided method can recover reasonable overall contours, it struggles to reconstruct detailed structures (\eg, the internal components of the airplane and watercraft) due to insufficient visual information. A potential solution is to incorporate multi-view images, leverage generative priors (\eg, diffusion models), or point cloud backbones~\cite{fei2026self} trained on large-scale datasets to enhance detail recovery.

A potential solution is to incorporate multi-view imagery and leverage generative priors, such as diffusion models, or large-scale pre-trained point cloud backbones~\cite{yang2025swin3d,fei2026self,ptv3} to enhance geometric detail recovery.

\section{Conclusion} 
\label{sec:conclusions}
In this work, we introduced MAGE, a novel view-guided point cloud completion framework that achieves efficient modality alignment and adaptive geometry enhancement. By projecting point cloud and image features into a shared latent space through a unified ViT backbone and cross-modality reconstruction supervision, MAGE effectively bridges the 2D–3D representation gap. Furthermore, the geometry-perceptive anchor point refinement branch updates the locations of anchor points extracted from incomplete inputs via global shape feature guidance and supervised training. More importantly, the adaptive geometry-aware self-attention module jointly captures local 3D structures and global contextual relationships and fuses the initial feature and attentive feature in a spatially variant manner. Extensive quantitative and qualitative evaluations confirm the superior completion performance of our approach. 

In the future, we would like to investigate its integration with other multi-modal generative frameworks. We are very interested in exploring similar design principles of our adaptive geometry-aware self-attention for cross-attention and ViT. We also plan to extend MAGE to more challenging scenarios, such as large-scale scene-level completion. 

\section{Declarations} 

\subsection*{Availability of data and materials}
1. ShapeNet-ViPC dataset can be accessed at \url{https://github.com/Hydrogenion/ViPC}. \\
2. Kitti cars dataset can be accessed at \url{https://github.com/yuxumin/PoinTr/blob/master/DATASET.md}. 

\subsection*{Competing interests}

The authors have no competing interests to declare that are relevant to the
content of this article.\\

\subsection*{Funding}
1. the National Natural Science Foundation of China (12494550, 12494553, and 12494554); \\
2. the Strategic Priority Research Program of the Chinese Academy of Sciences (XDB0640000 and
XDB0640200); \\
3. the Guangdong Basic and Applied Basic Research Foundation (2023B1515120026).

\subsection*{Authors' contributions}
Weize Quan: Conceptualization, Methodology, Validation, Visualization, Writing – original draft, Writing – review \& editing; Zhengwei Wu: Data curation, Validation, Visualization; Kai Wang: Conceptualization, Project administration, Writing – review \& editing; Dong-Ming Yan: Conceptualization, Project administration, Writing – review \& editing.

\subsection*{Acknowledgements}
This work was partially supported by the Strategic Priority Research Program of the Chinese Academy of Sciences (XDB0640000 and
XDB0640200); the National Natural Science Foundation of China (12494550 and 12494554); and the Guangdong Basic and Applied Basic Research Foundation (2023B1515120026).

\bibliographystyle{CVMbib}
\bibliography{acmart}

\end{document}